\documentclass[11pt]{article}

\usepackage[preprint]{acl} 

\usepackage[T1]{fontenc}
\usepackage[utf8]{inputenc}
\usepackage{microtype}
\usepackage{textcomp} 

\usepackage{amsmath}
\usepackage{subcaption} 
\usepackage{tikz}
\usepackage{booktabs}
\usetikzlibrary{arrows.meta, positioning, calc, fit, backgrounds, decorations.pathreplacing}

\title{Production and Perception in LLMs: A Token Probability Approach}

\author{
  Anna Marklová$^1$ \quad
  Jiří Milička$^1$ \quad
  Martina Vokáčová$^1$ \quad
  Rudolf Rosa$^2$ \\
  $^1$Faculty of Arts, Charles University, Prague \\
  $^2$Faculty of Mathematics and Physics, \\ Charles University, Prague \\
  \texttt{\{anna.marklova, jiri.milicka, martina.vokacova\}@ff.cuni.cz} \\
  \texttt{rosa@ufal.mff.cuni.cz}
}

\begin{document}

\maketitle

\begin{abstract}
The asymmetry between language production and perception has been well-documented in psycholinguistics. Whether large language models (LLMs) exhibit a functionally analogous distinction remains an open question, particularly given that LLMs rely on the same underlying mechanism—next-token prediction—for both input and output processing. In this exploratory study, we operationalize the production--perception distinction through direct token probability measurements rather than metalinguistic prompting. Using the base Llama-3.1-8B model, we generated poems under a production prompt and re-scored the same tokens under both rephrased production prompts and perception-oriented prompts. Across an extended experiment with four production and three perception prompts, production--perception distances consistently and substantially exceeded production--production distances, with non-overlapping ranges across conditions and an overall average ratio of approximately 1.8×. Near-ceiling correlations in the production--production control confirm that the effect is specific to communicative framing rather than prompt surface variation, and we show the effect replicates across five open-weight models (Llama-3.1-8B, EuroLLM-9B, gemma-2-9b-it, Mistral-7B-Instruct-v0.3, and Qwen2.5-7B-Instruct), spanning both base and instruction-tuned variants. Temporal analysis revealed that the perception prompt exerts its strongest influence at the beginning of the sequence, with divergence decaying as generated context accumulates, though the specific shape of this decay varies across prompt pairs. These findings suggest that prompt framing alone induces a production--perception distinction in LLM probability distributions, even within a decoder-only architecture.
\end{abstract}

\section{Introduction}
Human communication is not a mirror-image process in which the speaker and the hearer perform the same steps in reverse order. 
When we produce language, we select and encode a message with a particular communicative intention in mind. When we perceive language, we do not simply decode that message---we actively infer the most relevant interpretation given the context. Relevance Theory \citep{sperber1986relevance} models communication as ostensive-inferential: an utterance provides evidence of the speaker's intention and carries a presumption of optimal relevance; the addressee, guided by the expectation of such relevance, infers the speaker's intended meaning. 

When we try to apply this framework to large language models (LLMs), an immediate difficulty arises. Unlike humans, LLMs do not have communicative intentions in the traditional sense, nor do they perform inference of the user's intended meaning during interpretation. Instead, they process both input and output through the same underlying mechanism---next-token prediction. 
This symmetry is not merely a design choice but reflects a broader architectural shift in the field: whereas earlier encoder-decoder models \citep[e.g.,][]{vaswani2023attentionneed} maintained structurally distinct pathways for input processing and output generation---which could in principle support separate representations of perception and production---contemporary LLMs are predominantly decoder-only \citep{Radford2019GPT2}.
This architectural symmetry would seem to preclude any meaningful production--perception distinction. However, the framing provided by a prompt---whether it positions the model as a producer or an interpreter of text---may still shift the model's probability distributions in ways that mirror, at least functionally, the asymmetry outlined by Relevance Theory and other models and theories of human communication that describe the difference between speech production and perception\footnote{From here onward, we use the term \textit{perception} for the process most commonly used in the psycholinguistic literature to describe the role of the receiver in communication. Other terms, such as \textit{interpretation} or \textit{comprehension}, have also been used for this process.} (see, e.g., Cooperative Principle by \citealt{grice1975logic}, or the Common Ground framework by \citealt{clark1991grounding}).

Although LLMs are not cognitive agents in the human sense---that is, the intentions, goals, or agency attributed to them through anthropomorphisation are not inherent to their architecture---describing them as entirely devoid of agency can still be misleading. Their behavior is shaped by externally imposed objectives, constraints, and alignment procedures. In particular, instruction tuning and reinforcement learning from human feedback (RLHF) introduce normative pressures---such as helpfulness---that produce intention-like behavior. It creates an externally imposed goal structure that makes outputs appear as if communicative intentions were present. Goals should not be attributed to the model itself, but rather to personas that are simulated on that model, since if the simulated persona has intentions or goals, the intentions and goals are simulated as well \citep{shanahan2023role}.

Despite this apparent intentionality, the relationship between production and perception differs from human communication. \citet{Cuskley2024LLMLimitations} argue that in humans, production and perception are distinct developmental trajectories that interact but don't always track each other, shaped by multi-modal, embodied experience. In LLMs, by contrast, input processing and output generation rely on the same underlying mechanisms, making production essentially a conditioned continuation of perceived text. Autoregressive LLMs use shared parameters for text processing and generation \citep{Radford2019GPT2, LDR748}, resulting in a much stronger symmetry between production and perception than assumed in classical psycholinguistic models (see, e.g., \citealt{Levelt1989Speaking} for production, \citealt{MarslenWilsonTyler1980} for spoken language understanding, \citealt{Hendriks2014Asymmetries} for acquisition-based evidence of production--comprehension asymmetries, and \citealt{ArvidssonEtAl2024Conversational} for conversational neuroimaging evidence).

Ways of measuring the difference between production and perception in LLMs are limited. Most studies on LLMs so far examined either production capabilities---e.g., on poetry \citep{porter2024poetry}, multi-genre texts \citep{milička2025benchmarkstylisticvariationllmgenerated}---or perception capabilities---e.g., about garden-path sentences \citep{Futrell2019NeuralLM}, pragmatic resolution \citep{Hu2023PragmaticComparison}, or acceptability judgments \citep{Warstadt2020BLiMP}. Some researchers tested the distinction between production and perception in LLMs using behavioral tasks. \citet{lam-etal-2025-leveraging} examined the phenomenon of implicit causality verbs, where humans are known to show an asymmetry: they are more subject-biased when interpreting an ambiguous pronoun than when freely producing a continuation. As discovered, LLMs can only partially and inconsistently replicate this human asymmetry. Larger models (LLaMA-70B, GPT-4o) do show some human-like distinction between the two tasks under specific prompting conditions, but the effect is weaker in magnitude than in humans, highly dependent on how the question is framed, and often disappears or even reverses with different prompts or smaller models. The authors conclude that LLMs generally struggle to recognize 

the difference between production and perception the way humans do, and, similarly to \citet{Cuskley2024LLMLimitations}, the authors conclude that it is because LLMs use the same underlying mechanism for both input and output, and the human-like asymmetry between the two processes is not naturally encoded in them. 

In our exploratory study, we propose using token probability distributions to measure the distinction between production and perception. Our approach is motivated by \citet{hu-levy-2023-prompting}, who demonstrate that eliciting metalinguistic judgments through prompting is not a reliable substitute for direct probability measurements in LLMs. They show that metalinguistic responses, i.e., model outputs elicited by asking the model to judge or evaluate linguistic content, systematically diverge from the model's underlying token probability distributions, and that this divergence increases as the prompt moves further from a direct next-word prediction task. This finding supports our decision to measure the production--perception distinction through token probabilities directly, rather than through prompted evaluative responses.

\section{Methodology}

\subsection{Experimental Setup and Hypotheses}
The core logic of our design lies in the following premise: if switching from a production prompt to a perception prompt merely rephrases the same communicative role, the model's token probability distributions should remain largely stable. If, however, the change in framing genuinely shifts the model into a functionally different mode---that of a perceiver rather than a producer---the probability distributions should diverge systematically. To test this, we used a simple production prompt (e.g., "Please write a poem about a rabbit") to elicit a poem, and recorded token probabilities during generation.

The same poem was then re-evaluated token by token under two alternative prompt framings: a rephrased production prompt (e.g., "I would like you to write a poem about a rabbit"), serving as a control condition (pro\_pro), and a perception-oriented prompt (e.g., "Please rate this poem about a rabbit"), constituting the experimental condition (pro\_perc). We chose this design to keep the two prompts as similar as possible, isolating communicative framing as the main variable. The whole procedure is visualized in Fig. \ref{fig:schema}.

\begin{figure*}[ht]
\centering
\resizebox{\textwidth}{!}{%
\begin{tikzpicture}[
    >=Stealth,
    font=\small,
    prompt/.style={draw, rounded corners=3pt,
                   inner sep=5pt,
                   text width=4.8cm,
                   align=left, font=\small},
    modelbox/.style={draw, rounded corners=2pt, fill=gray!12,
                     minimum width=1cm, minimum height=0.5cm,
                     font=\small\bfseries},
    tok/.style={draw, fill=white, minimum width=0.85cm, minimum height=0.5cm,
                inner sep=2pt, font=\scriptsize\ttfamily}, 
    pval/.style={font=\scriptsize, align=center},
]

\def\valX{8.8}      
\def\valS{0.92}     

\def\llmX{6.6}
\def\outX{7.6}

\node[font=\normalsize\bfseries, anchor=north west] at (0, 0.7) {Step 1: Generate};

\node[font=\footnotesize\bfseries, anchor=south west] at (\valX, 0.65)
    {Example (hedgehog poem, first 7 tokens):};

\node[prompt, fill=white, anchor=north west] (prodP) at (0, 0)
    {\textbf{Primary production prompt}\\[1pt]
     \footnotesize\textit{``Please write a poem}\\
     \footnotesize\textit{about a hedgehog.''}};

\node[modelbox] (llm1) at (\llmX, -0.5) {LLM};
\draw[->, thick, black!70] (prodP.east) -- (llm1.west);
\node at (\outX, -0.5) {$\longrightarrow$};

\node[font=\tiny, anchor=south] at ({\valX + 3*\valS}, 0.0)
    {generated tokens $t_1, t_2, \ldots, t_n$};

\foreach \tok/\i in {In/0, the/1, garden/2, {,}/3, I/4, saw/5, a/6} {
    \node[tok] at ({\valX + \i*\valS}, -0.5) {\tok};
}
\node[font=\small] at ({\valX + 7*\valS}, -0.5) {$\cdots$};

\node[font=\scriptsize\bfseries, green!50!black, anchor=east]
    at ({\valX - 0.1}, -1.2) {$p_0$: };
\foreach \val/\i in {.007/0, .579/1, .178/2, .431/3, .050/4, .355/5, .887/6} {
    \node[pval, green!50!black] at ({\valX + \i*\valS}, -1.2) {\val};
}
\node[font=\scriptsize, green!50!black] at ({\valX + 7*\valS}, -1.2) {$\cdots$};

\node[font=\normalsize\bfseries, anchor=north west] at (0, -2.0)
    {Step 2: Score each token under both prompts};

\node[prompt, fill=red!6, inner ysep=7pt, anchor=north west] (ctxA) at (0, -2.6)
    {\textbf{Prompt~1} (alternative production)\\[1pt]
     \footnotesize\textit{``I would like you to write}\\
     \footnotesize\textit{a poem about a hedgehog.''}\\
     \footnotesize + $t_1, \ldots, t_{i-1}$};

\node[modelbox] (llm2a) at (\llmX, -3.5) {LLM};
\draw[->, thick, red!70!black] (ctxA.east) -- (llm2a.west);
\node at (\outX, -3.5) {$\longrightarrow$};

\node[font=\scriptsize\bfseries, red!70!black, anchor=east]
    at ({\valX - 0.1}, -3.5) {$p_1$: };
\foreach \val/\i in {.008/0, .571/1, .183/2, .428/3, .051/4, .361/5, .884/6} {
    \node[pval, red!70!black] at ({\valX + \i*\valS}, -3.5) {\val};
}
\node[font=\scriptsize, red!70!black] at ({\valX + 7*\valS}, -3.5) {$\cdots$};

\node[prompt, fill=blue!6, anchor=north west] (ctxB) at (0, -4.7)
    {\textbf{Prompt~2} (perception)\\[1pt]
     \footnotesize\textit{``Here is a poem about a}\\
     \footnotesize\textit{hedgehog, please rate it.''}\\[-1pt]
     \footnotesize + $t_1, \ldots, t_{i-1}$};

\node[modelbox] (llm2b) at (\llmX, -5.6) {LLM};
\draw[->, thick, blue!70!black] (ctxB.east) -- (llm2b.west);
\node at (\outX, -5.6) {$\longrightarrow$};

\node[font=\scriptsize\bfseries, blue!70!black, anchor=east]
    at ({\valX - 0.1}, -5.6) {$p_2$: };
\foreach \val/\i in {.005/0, .287/1, .094/2, .225/3, .076/4, .283/5, .799/6} {
    \node[pval, blue!70!black] at ({\valX + \i*\valS}, -5.6) {\val};
}
\node[font=\scriptsize, blue!70!black] at ({\valX + 7*\valS}, -5.6) {$\cdots$};

\pgfmathsetmacro{\brX}{\valX + 7.7*\valS}

\draw[decorate, decoration={brace, amplitude=4pt}, thick]
    (\brX, -1.0) -- (\brX, -3.7)
    node[midway, right=8pt, font=\scriptsize]
    {\rotatebox{-90}{%
      \begin{tabular}{c}
      compare $(p_{0,i},\, p_{1,i})$\\
      for each position $i$
      \end{tabular}}};

\draw[decorate, decoration={brace, amplitude=4pt}, thick]
    ({\brX + 1.4}, -1.0) -- ({\brX + 1.4}, -5.8)
    node[midway, right=8pt, font=\scriptsize]
    {\rotatebox{-90}{%
      \begin{tabular}{c}
      compare $(p_{0,i},\, p_{2,i})$\\
      for each position $i$
      \end{tabular}}};

\end{tikzpicture}%
}
\caption{Schematic of the experimental procedure.
         \textbf{Step~1:} A poem is generated under the production prompt;
         $p_0$ is the probability assigned to each token during generation.
         \textbf{Step~2:} The same tokens are scored under a rephrased
         production prompt ($p_1$) and a perception prompt ($p_2$).
         The two comparisons---$(p_0, p_1)$ and $(p_0, p_2)$---correspond
         to the \textsc{prod-prod} and \textsc{prod-perc} conditions.
         Values shown are from the first seven tokens of the hedgehog poem
         ($p_1$ values are illustrative).}
\label{fig:schema}
\end{figure*}

Since LLMs rely on the same underlying representations for both input processing and output generation, we might expect the two conditions to yield similar probability distributions. However, if a systematic asymmetry is observed---specifically, if token probabilities diverge more strongly in the pro\_perc condition than in the pro\_pro control---this would suggest that prompt framing induces a functionally meaningful distinction at the level of token probabilities, analogous to the production–perception asymmetry documented in human language processing.

\subsection{Model and stimuli}

We used the self-hosted \texttt{Llama-3.1-8B} by Meta as the target model to generate both the poem and the probabilities. The choice of model was motivated by earlier studies suggesting that smaller LMs better mimic human cognitive behavior during some tasks than larger ones \citep{shain2024logarithmic, oh2023surprisal, kuribayashi2022context}. We chose the base model to avoid an implicit "helpful assistant" effect, often present in fine-tuned models, and to ensure that the communicative role is invoked by the prompt itself.

We generated poems about 1{,}089 topics (individual animals and animal pairs,
e.g.\ \emph{a hedgehog}, \emph{a tiger and a penguin} etc.) using a primary
production prompt (``please write a poem about~\ldots''), sampling 100 tokens at temperature = 1. The probabilities $p_0$ of the sampled tokens are recorded.

Secondary probabilities were obtained for these two scenarios:

\begin{itemize}
    \item \textbf{Production--Production (\textsc{prod-prod}):}
For each generated poem, we make second forward pass, starting with secondary
production prompt (``I would like you to write a poem about~\ldots''), and then feeding the same generated token IDs back into the model one at a time to obtain $p_1$.
    \item \textbf{Production--Perception (\textsc{prod-perc}):}
The third forward pass starts with perception prompt (``Here is a poem about {\ldots}, please rate the poem.''), and then again feeding the same generated token IDs back into the model one at a time to obtain $p_2$. We did not let the model generate the rating since it is irrelevant for our study; the perception prompt serves solely as a framing context within which the probabilities of the poem tokens are computed.

\end{itemize}

In all three cases, the assistant--user environment was simulated, since the base model did not tend to generate a poem at all, unless triggered by \emph{user} / \emph{assistant} marking. 

For technical reasons (to make the procedure easily parallelized on multiple machines), the poems and $p_0$'s were calculated twice for both scenario, but this should not influence results whatsoever. This procedure yields approximately 93{,}000 token-level $(p_1, p_2)$ pairs per condition (93{,}697 for \textsc{prod-perc}; 93{,}341 for \textsc{prod-prod}).

\subsection{Measures}

We quantified the divergence between the two probability streams at two levels of granularity:

\begin{enumerate}
    \item \textbf{Per-poem Pearson correlation} between $p_0$ and $p_1$, and between $p_0$ and $p_2$,
          capturing the linear agreement of the two probability profiles.
    \item \textbf{Per-poem mean absolute distance}
          $\overline{|p_0 - p_1|}$, and $\overline{|p_0 - p_2|}$, capturing the magnitude of token-level shifts
          independently of co-variation.
\end{enumerate}

\noindent
To examine how the prompt-framing effect evolves over the course of generation,
we also computed position-wise mean distance for each of the first 100 token
positions and  modeled the position-wise means over positions 10--100
(excluding the high-variance boundary region immediately following the prompt). The model we have chosen is an exponential decay with asymptote,
$y = a\,e^{-bx} + c$,
where $a$ captures the initial excess divergence, $b$ the decay rate,
and $c$ the irreducible residual divergence that persists regardless of
sequence length. This functional form is motivated by the assumption that each additional
token of context reduces the remaining prompt influence by a roughly
constant fraction (multiplicative dilution), analogous to geometric mixing
in ergodic processes.

All confidence intervals and fits were estimated via bootstrap resampling
(10{,}000 iterations, resampling poems).

\section{Results}

\subsection{Global token-level agreement}

\begin{figure}[h]
    \centering
    \begin{subfigure}[t]{0.48\columnwidth}
        \includegraphics[width=\columnwidth]{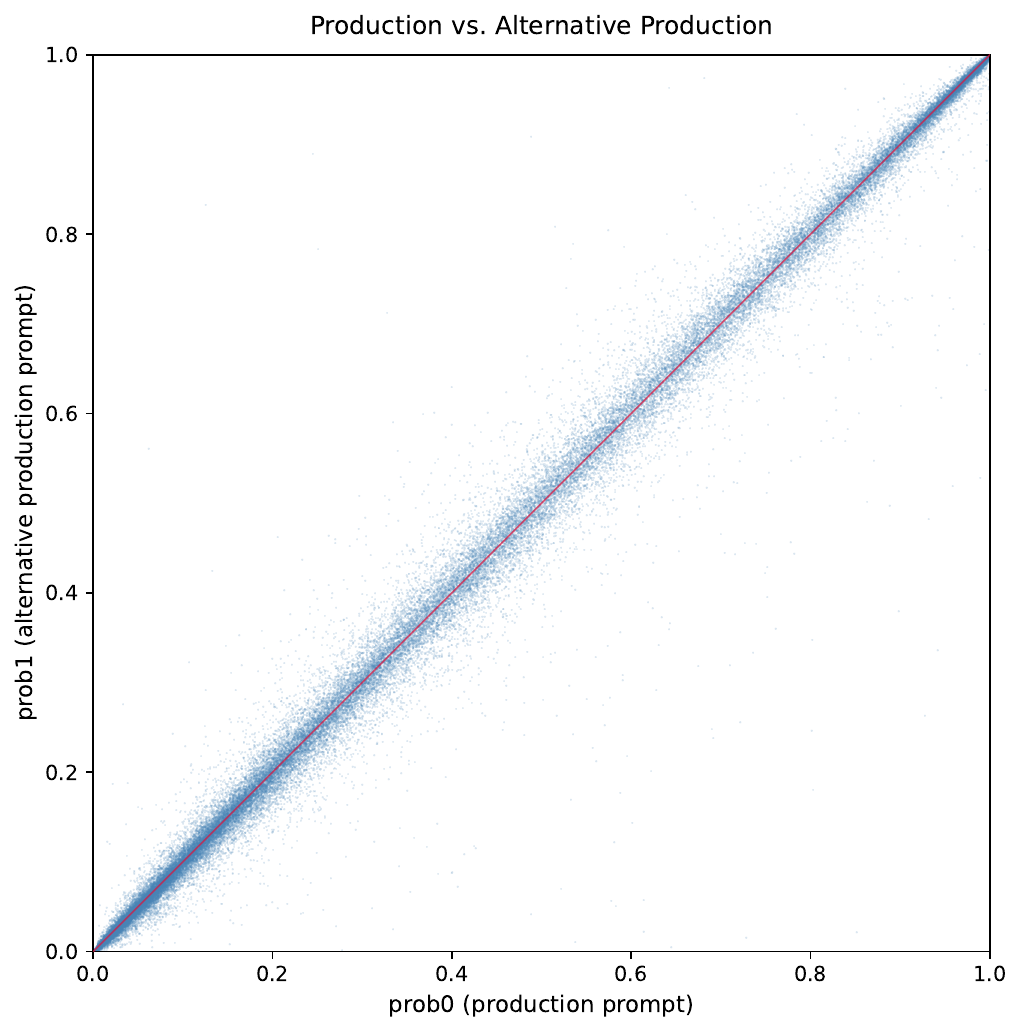}
        \caption{\textsc{prod-prod}}
        \label{fig:scatter_pro}
    \end{subfigure}
    \hfill
    \begin{subfigure}[t]{0.48\columnwidth}
        \includegraphics[width=\columnwidth]{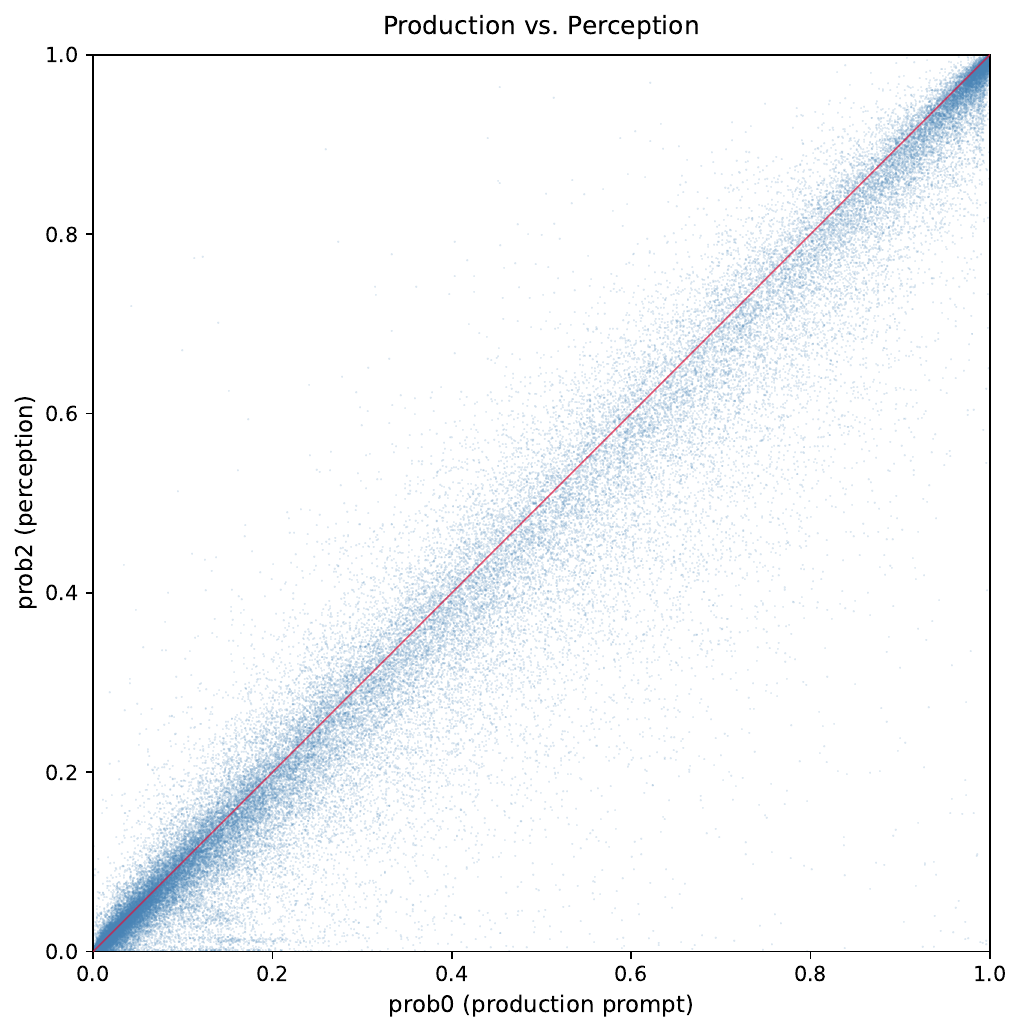}
        \caption{\textsc{prod-perc}}
        \label{fig:scatter_perc}
    \end{subfigure}
    \caption{Token-level probability scatter plots.
             Each point is one token; the red line marks perfect agreement.}
    \label{fig:scatter}
\end{figure}

Figure~\ref{fig:scatter} plots $p_0$ against $p_1$ and $p_2$ for all tokens.
In \textsc{prod-prod}, points cluster tightly along the diagonal, indicating that
prompt rephrasing alone has negligible effect on token probabilities.
In \textsc{prod-perc}, scatter around the diagonal is visibly larger,
reflecting a systematic divergence introduced by the change in communicative framing.

\subsection{Per-poem distances}
\label{sec:results:dist}
\begin{figure}[h]
    \centering
    \includegraphics[width=\columnwidth]{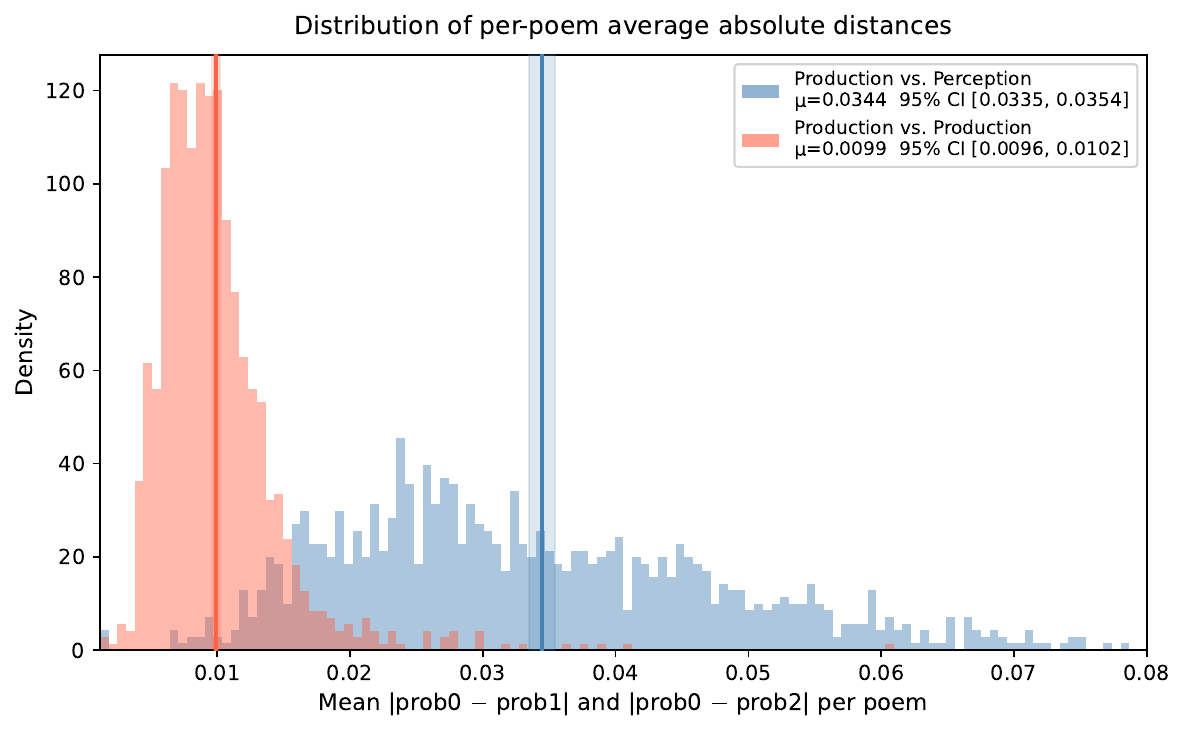}
    \caption{Distribution of per-poem mean distances across ${\sim}1{,}089$~poems.
             Vertical lines: group means; shaded bands: 95\% bootstrap CIs.}
    \label{fig:distributions}
\end{figure}

Figure~\ref{fig:distributions} summarises the per-poem statistics.
The mean absolute distance in the \textsc{prod-perc} condition
($\mu = 0.034$, $\mathit{SD} = 0.017$) is approximately
3.5$\times$ larger than in \textsc{prod-prod}
($\mu = 0.010$, $\mathit{SD} = 0.005$), with non-overlapping 95\% bootstrap CIs.

\subsection{Temporal dynamics}

\begin{figure}[h]
    \centering
    \begin{subfigure}[t]{\columnwidth}
        \includegraphics[width=\columnwidth]{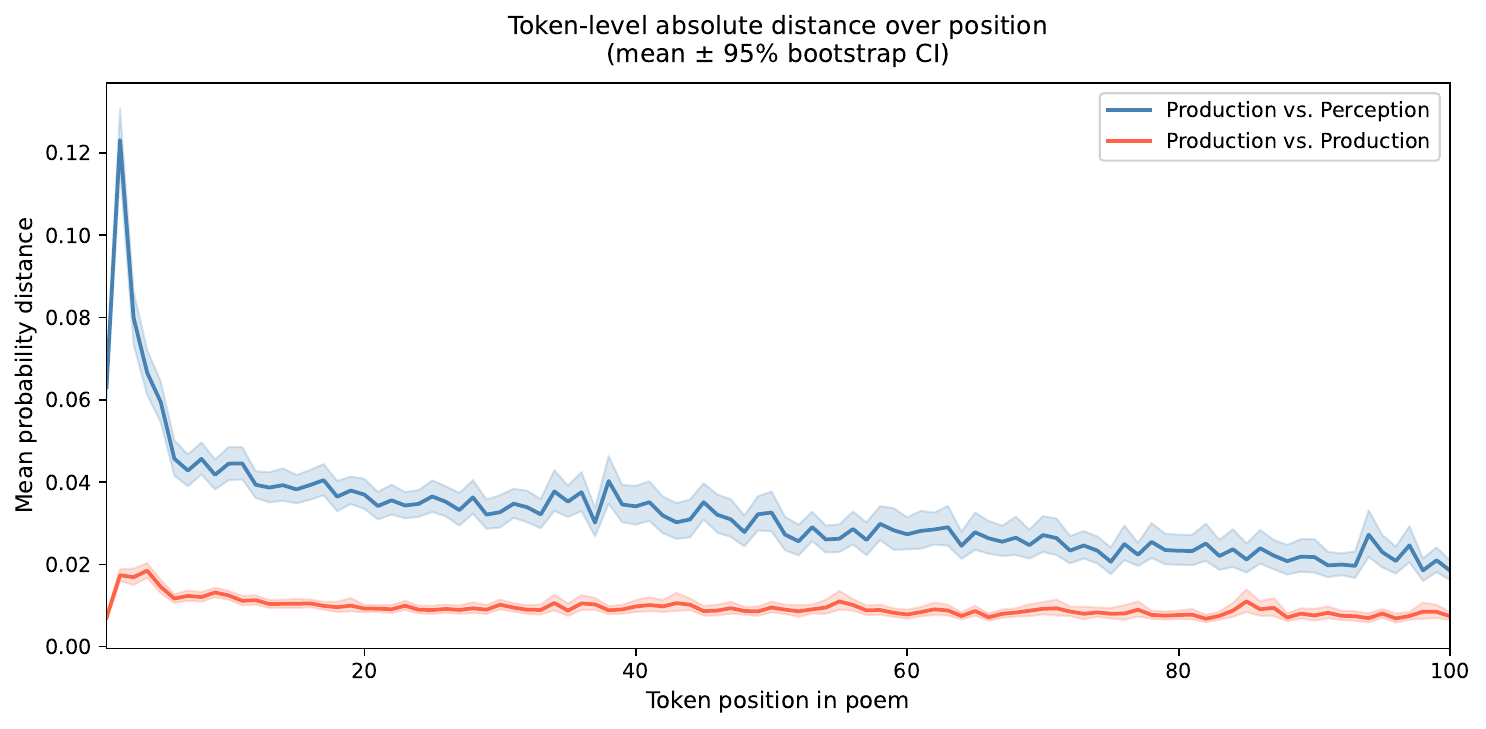}
        \caption{Position-wise mean distances with 95\% bootstrap CIs.}
        \label{fig:time_raw}
    \end{subfigure}
    \\
    \begin{subfigure}[t]{\columnwidth}
        \includegraphics[width=\columnwidth]{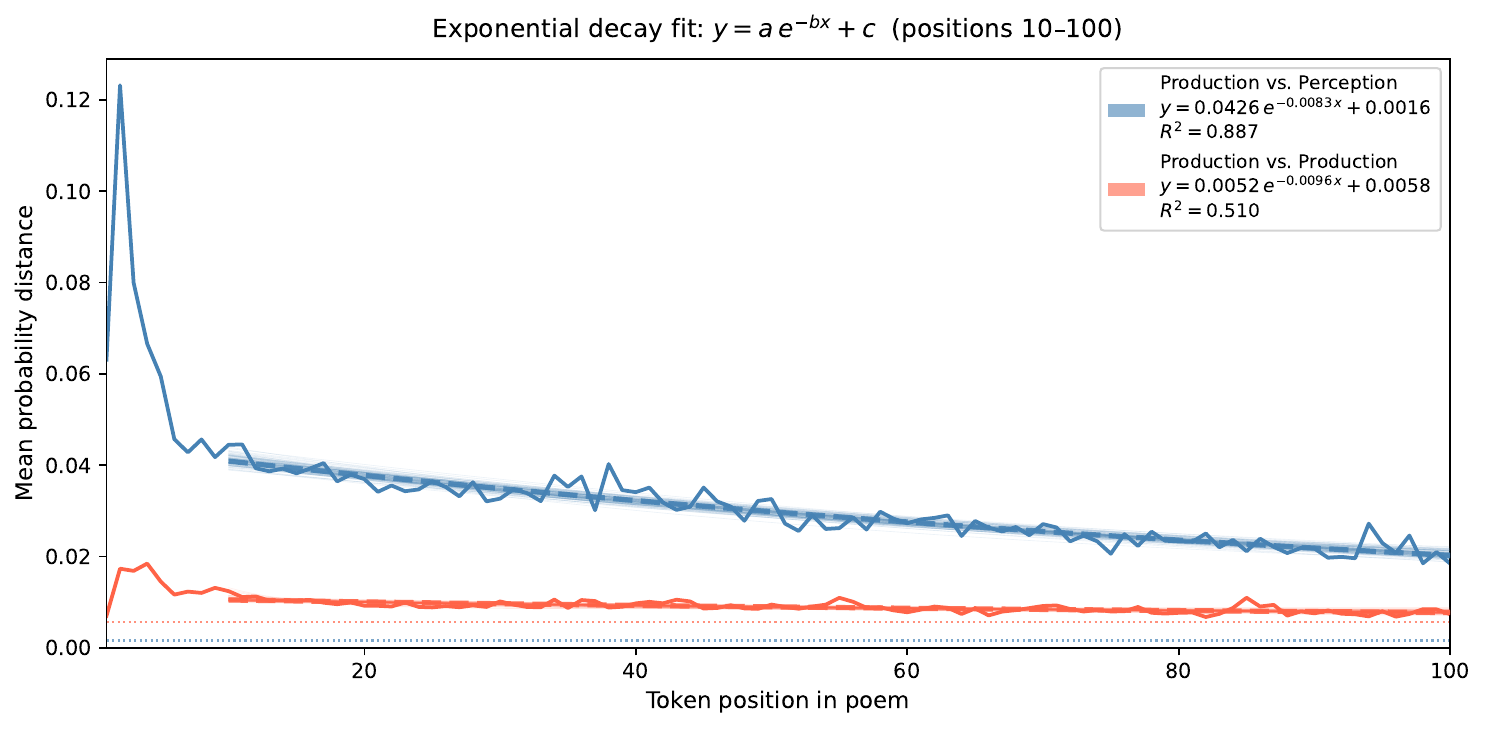}
        \caption{Bootstrap exponential decay fits (positions 10--100).}
        \label{fig:time_fit}
    \end{subfigure}
    \caption{Temporal dynamics of the production--perception divergence.}
    \label{fig:time}
\end{figure}

Figure~\ref{fig:time} shows how distances ($|p_0 - p_1|$ and $|p_0 - p_2|$) evolve over token position.
Both conditions exhibit a decreasing trend: as generated context accumulates,
the local context increasingly dominates over the prompt signal, and the two
probability seems to converge.

The separation between conditions persists throughout the sequence and remains
outside the respective 95\% CIs, but as the exponential decay model suggests, at some point in the future, it is perfectly possible that the lines will meet and that the production--perception distance can be even smaller than the production--production distances in the long run.

Bootstrap exponential decay fits over positions 10--100 yield:
\begin{align*}
    \textsc{prod-perc}{:} \quad & y = 0.0427\, e^{-0.0083\,x} + 0.0016 \\
    \textsc{prod-prod}{:} \quad & y = 0.0052\, e^{-0.0096\,x} + 0.0058
\end{align*}
To be more precise, the goodness of fit, as measured by $R^2$ for \textsc{prod-perc} is $0.8869$ and the parameters of the model and their  95 \% confidence intervals are 
    $a = 0.04265$  [CI: $0.02985$, $0.04625$], 
    $b = 0.00829$  [CI: $0.00705$, $0.01811$], 
    $c = 0.00164$  [CI: $0.00000$, $0.01651$].

The $R^2$ for \textsc{prod-prod} is  $0.5100$ and the parameters are 
    $a = 0.00523$  [CI: $0.00345$, $0.01136$], 
    $b = 0.00960$  [CI: $0.00286$, $0.09189$],
    $c = 0.00576$  [CI: $0.00000$, $0.00851$].

The total initial divergence ($a+c$) in \textsc{prod-perc} is roughly $4\times$ that of \textsc{prod-prod}, confirming a stark contrast when the communicative framing changes. More notably, the context-sensitive portion of this divergence (parameter $a$) is ${\sim}8\times$    larger in \textsc{prod-perc} ($a \approx 0.0427$ vs.\ $0.0052$). 

While the relative rate of decay is comparable across both conditions ($b \approx 0.0083$ vs.\ $0.0096$), the massive initial magnitude in \textsc{prod-perc} results in a much steeper absolute drop early in the text. This exponential trajectory demonstrates that the perception framing exerts a powerful initial influence that is rapidly and progressively overridden by accumulated context, eventually asymptoting to a near-zero baseline ($c \approx 0.0016$).

\section{Results for more prompts}\label{sec:results:extended}

The results reported in the previous section are based on a single production prompt
and a single perception prompt, leaving open the possibility that the observed
divergence is tied to the specific wording of those prompts rather than to the
production--perception distinction per se.  To assess robustness, we replicated
the experiment with a set of four production prompts (\textit{prod$_0$--prod$_3$})
and three perception prompts (\textit{perc$_0$--perc$_2$}), yielding 12
production--perception combinations and 12 production--production combinations
(one for each off-diagonal pair from the same set of four prompts).
All other parameters were held constant: the same 1{,}089 topics, the same model and hyperparameters,
and the same 100-token generation limit.

\subsection{Prompt variants}

Table~\ref{tab:prompts} lists all prompt templates used in the extended
experiment.  The production prompts span a range of registers, from a minimal
direct request (\textit{prod$_0$}) to longer, more contextualised formulations
(\textit{prod$_2$}, \textit{prod$_3$}). The perception prompts likewise vary in how
explicitly they position the model as an evaluator, from a brief rating
instruction (\textit{perc$_0$}) to a more discursive opinion framing (\textit{perc$_2$}).
In all conditions the assistant--user turn structure is preserved, as in the
original experiment.

\begin{table}[h]
\centering
\small
\begin{tabular}{lp{6.2cm}}
\toprule
\textbf{ID} & \textbf{Prompt} \\
\midrule\noalign{\vspace{2pt}}
\multicolumn{2}{l}{\textit{Production prompts}} \\[2pt]
\textit{prod$_0$} & ``Please write a poem about \textit{X}.'' \\
\textit{prod$_1$} & ``I would like you to write a poem about \textit{X}.'' \\
\textit{prod$_2$} & ``Hi, my kid wants to listen to a poem about \textit{X},
                 would you be so kind and create one?'' \\
\textit{prod$_3$} & ``I really love poetry but I have never seen a poem about
                 \textit{X}, compose one, please.'' \\\noalign{\vspace{4pt}}
\multicolumn{2}{l}{\textit{Perception prompts}} \\[3pt]
\textit{perc$_0$} & ``Here is a poem about \textit{X}, please rate the poem.'' \\
\textit{perc$_1$} & ``I would like to hear your opinion on this poem about
                  \textit{X}:'' \\
\textit{perc$_2$} & ``I have read a poem about \textit{X} and I would like to
                  know what you think about it.  Here is the poem:'' \\
\bottomrule
\end{tabular}
\caption{Production and perception prompt templates used in the extended
         experiment.  \textit{X} denotes the poem topic (animal or animal
         pair).  In the \textsc{prod-prod} condition the same four production
         prompts serve as both the primary and the secondary prompt (in combination).}
\label{tab:prompts}
\end{table}

\subsection{Overall statistics}
\begin{figure}[h]
    \centering
    \includegraphics[width=\columnwidth]{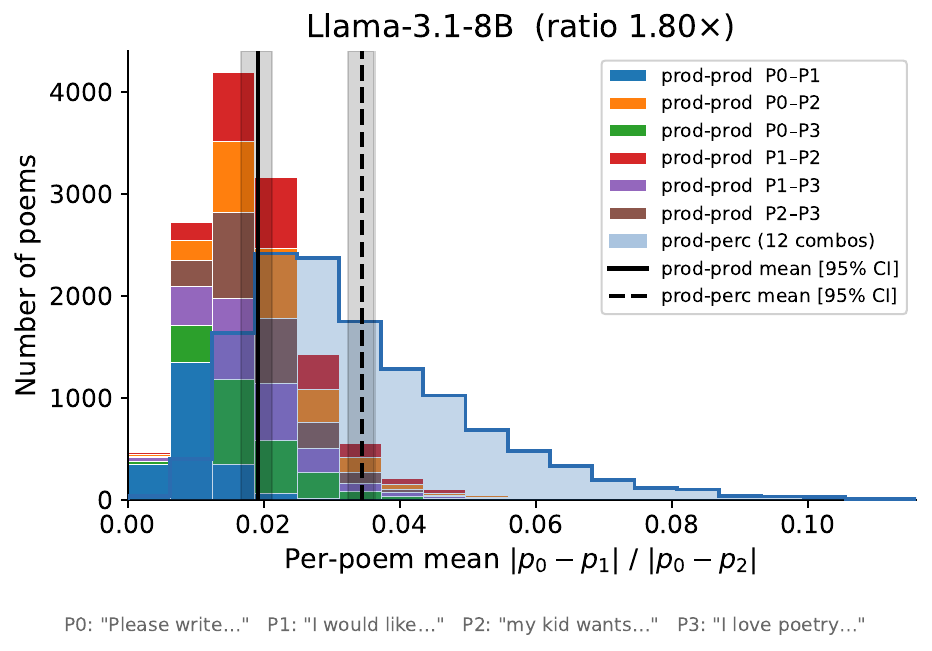}
    \caption{Per-poem distances for \texttt{Llama-3.1-8B}.  The \textsc{prod-prod}
             distances are decomposed by the six production-prompt pairs
             (one colour per pair; each pair and its reverse are pooled). The \textsc{prod-perc} distances (pale blue) are pooled together. Vertical lines mark the overall means with 95\%
             bootstrap CI bands.}
    \label{fig:scenario_dist}
\end{figure}

The extended results closely replicate the original finding.  In the
\textsc{prod-perc} condition, mean absolute distances range from $0.028$ to
$0.040$ across all 12 production--perception prompt combinations
(grand mean $\approx 0.035$), almost identical to the $0.034$ reported in
Section~\ref{sec:results:dist}.  Crucially, this range does not overlap with
the highest \textsc{prod-prod} distance ($0.024$). The mean difference between \textsc{prod-prod} and \textsc{prod-perc} is $0.016$ [95 \% CI: $0.013$, $0.019$] confirming that the
production--perception asymmetry is not an artefact of a particular prompt
pairing. Figure~\ref{fig:scenario_dist} shows the full per-poem
distributions for \texttt{Llama-3.1-8B}, decomposed by prompt pair: the
\textsc{prod-perc} mass (blue) sits clearly above the \textsc{prod-prod}
components (red), and the two overall means are separated with non-overlapping
95\% bootstrap confidence intervals.

The distinction between \textsc{prod-perc} and \textsc{prod-prod} is also visible in correlation indices and in the exponential decay model parameters (see Table~\ref{tab:scenario_results} in Appendix~\ref{app:scenario}). Among these parameters, we are primarily interested in $\hat{b}$, representing the rate of the curve's decay, and $\hat{c}$, representing the asymptote to which it would eventually drop. The results indicate that the behavior of these parameters is relatively chaotic; therefore, it cannot be simply concluded that any specific condition systematically exhibits a lower asymptote or a steeper decay curve.


\subsection{Generalization across models}
\label{sec:results:models}

The results reported so far are based on a single model. To test whether the production--perception asymmetry is specific to \texttt{Llama-3.1-8B} or a more general property, we re-ran the identical extended pipeline (the same 1{,}089 topics, the four production and three perception prompts, and the same 100-token generation limit) on four additional open-weight models: \texttt{EuroLLM-9B}, \texttt{gemma-2-9b-it}, \texttt{Mistral-7B-Instruct-v0.3}, and \texttt{Qwen2.5-7B-Instruct}. Of these, \texttt{EuroLLM-9B} is a base model like \texttt{Llama-3.1-8B}, whereas the remaining three are instruction-tuned; this lets us probe whether the effect survives instruction tuning and the associated ``helpful assistant'' framing. 

\begin{figure*}[t]
    \centering
    \includegraphics[width=0.89\textwidth]{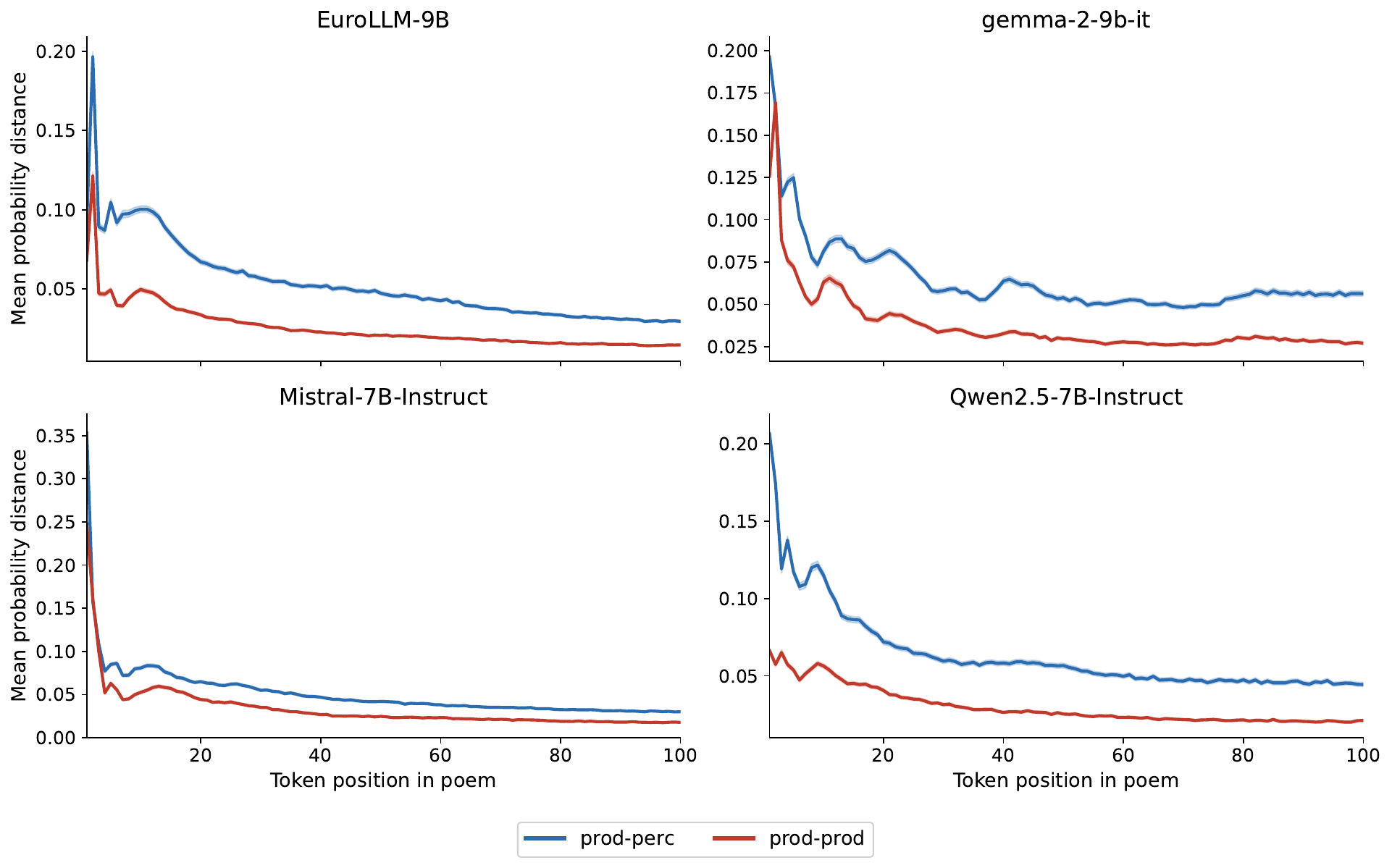}
    \caption{Token-level absolute distance over position for the four additional
             models (mean $\pm$ 95\% CI).}
    \label{fig:time_allmodels}
\end{figure*}

The asymmetry replicates in every model. As Figure~\ref{fig:time_allmodels} shows, the \textsc{prod-perc} curve lies above the \textsc{prod-prod} curve at every token position in all four additional models: the distance between the probabilities assigned under production and perception framings is consistently larger than the distance between two production framings. The effect is present in the instruction-tuned models as well as the base models, indicating that it is not an artefact of the absence of alignment. The temporal pattern is likewise preserved: in every model the \textsc{prod-perc} distance starts high at the beginning of the poem and decays towards the \textsc{prod-prod} curve as generated context accumulates. \texttt{Mistral-7B-Instruct} and \texttt{EuroLLM-9B} behave similarly to \texttt{Llama-3.1-8B}, but the pattern does not fully generalize: in \texttt{gemma-2-9b-it} and \texttt{Qwen2.5-7B-Instruct} the two curves do not converge, and the parameter~$c$ remains distinct.

Quantitatively, the effect is substantial and consistent across all five models.
The overall mean \textsc{prod-perc} distance exceeds the \textsc{prod-prod}
distance by a factor of $1.80$ [95\% CI $1.77$, $1.83$] for \texttt{Llama-3.1-8B},
$2.09$ [$2.07$, $2.10$] for \texttt{EuroLLM-9B},
$1.72$ [$1.71$, $1.74$] for \texttt{gemma-2-9b-it},
$1.53$ [$1.52$, $1.55$] for \texttt{Mistral-7B-Instruct}, and
$2.11$ [$2.10$, $2.13$] for \texttt{Qwen2.5-7B-Instruct};
every interval lies well above~$1$ (each ratio pools ${\sim}13{,}000$ poems per
condition), and in each model the two per-condition means are separated with
non-overlapping 95\% bootstrap CIs.

\begin{figure*}[t]
    \centering
    \includegraphics[width=0.89\textwidth]{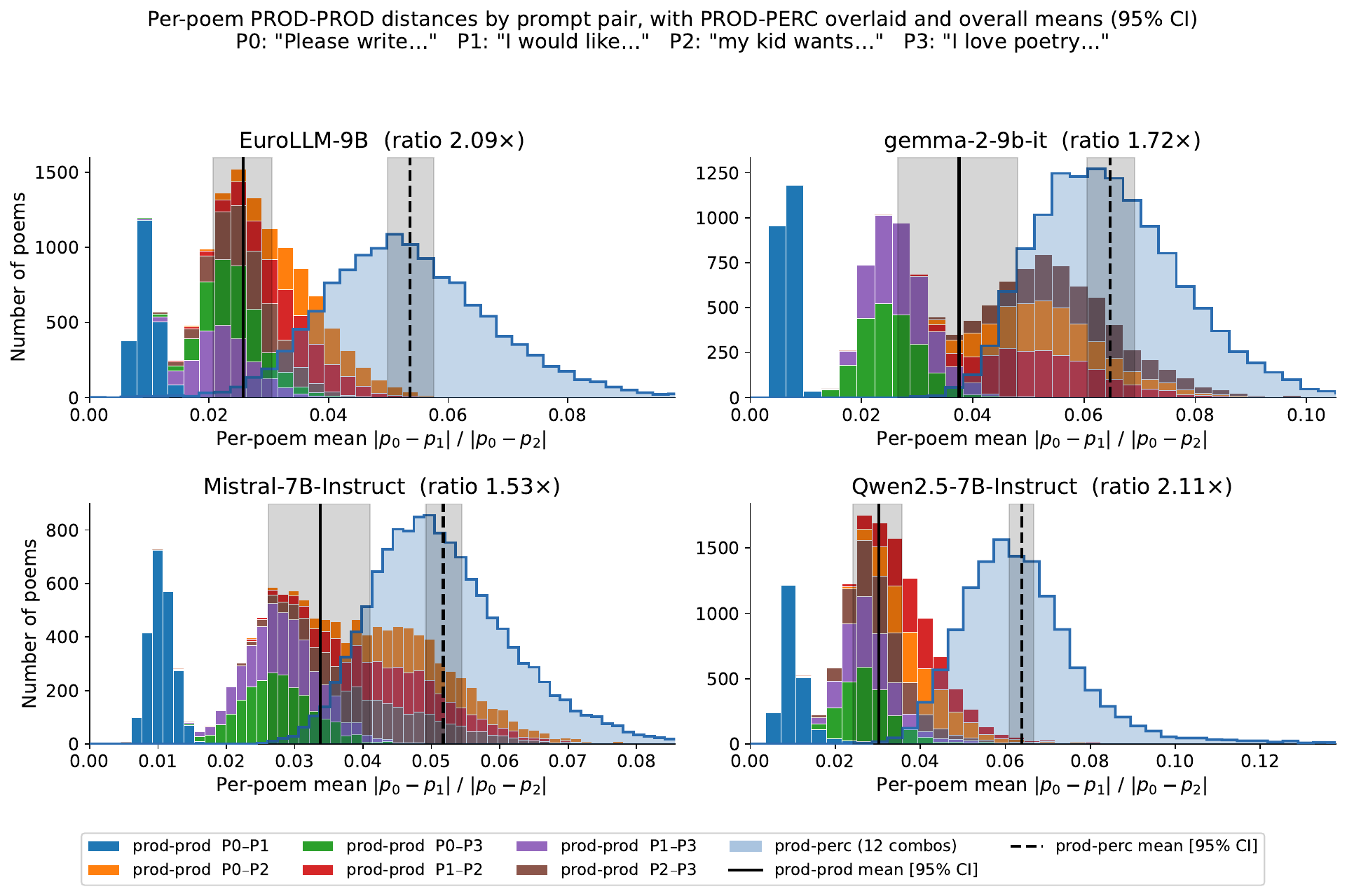}
    \caption{Per-poem distances for the four additional models.  \textsc{prod-prod}
             distances are decomposed by the six unordered production-prompt pairs
             (one colour per pair).
             Vertical lines mark the overall means (solid: \textsc{prod-prod},
             dashed: \textsc{prod-perc}) with 95\% bootstrap CIs; panel titles
             report the \textsc{prod-perc}/\textsc{prod-prod} ratio.}
    \label{fig:bypair_grid}
\end{figure*}

One apparent irregularity is worth addressing. When the per-poem
\textsc{prod-prod} distances are pooled across prompt pairs, their distribution
can appear bimodal (e.g.\ \texttt{EuroLLM-9B}) or trimodal
(e.g.\ \texttt{gemma-2-9b-it}) rather than unimodal. This is a pooling artefact
rather than a sign of an ill-behaved effect: some production prompt pairs are
near-paraphrases of one another (for instance \textit{prod$_0$}, ``Please write
a poem about \textit{X},'' and \textit{prod$_1$}, ``I would like you to write a
poem about \textit{X}''), and for most models these near-identical pairs yield
very small distances while more dissimilar pairs yield larger ones, producing
separate humps in the pooled histogram. Decomposing the \textsc{prod-prod}
distances by individual prompt pair shows that each pair is cleanly unimodal
(Figure~\ref{fig:bypair_grid}), confirming that the multimodality reflects the
mixture of qualitatively different prompt pairs rather than any instability in
the underlying measurement.

\section{Conclusion}
This exploratory study examined the production--perception distinction in LLMs through direct token probability measurements, examining whether prompt framing alone---with no change in the generated text---induces a systematic divergence in the model's probability distributions. Our results show that it does. Probabilities assigned to poem tokens under a perception-oriented prompt diverge substantially and consistently from those assigned under a rephrased production prompt, with the production--perception condition showing 1.82× larger mean absolute distances than the production--production control in \texttt{Llama-3.1-8B}. This divergence is not an artifact of prompt rephrasing per se, as the production--production condition yields near-ceiling correlations across poems, confirming that the effect is specific to the change in communicative framing. The extended experiment across four production and three perception prompts further strengthens this conclusion, confirming that the asymmetry is not an artifact of any particular prompt pairing but a robust consequence of shifting communicative framing (see Section~\ref{sec:results:extended}).

We also found that the effect changes over the course of the poem. The perception prompt has the biggest impact at the beginning of the sequence, but as more tokens are generated, the two conditions become more similar. In other words, the prompt matters most at the start, but the generated text gradually takes over and reduces the difference between conditions. This pattern is consistent across prompt pairs at the level of overall magnitude, though as noted in Section~\ref{sec:results:extended}, the specific shape of the temporal dynamics (decay rate and asymptote) varies considerably across prompt combinations and models.

These results are relevant for several reasons. First, they demonstrate that a base model without instruction tuning or fine-tuning encodes something functionally analogous to a production--perception distinction at the level of its probability distributions. This is noteworthy given that, as \citet{Cuskley2024LLMLimitations} and \citet{lam-etal-2025-leveraging} argue, LLMs use the same underlying mechanism for input and output processing. One might therefore expect no difference at all. Our results suggest, however, that some form of the distinction is nonetheless present, even if it may be weaker and more context-dependent than in humans. Moreover, this distinction is not idiosyncratic to a single model: it replicates across five open-weight models and persists in instruction-tuned models as well as base models (Section~\ref{sec:results:models}).
This finding is particularly noteworthy in light of the broader architectural trajectory of the field: the practical abandonment of encoder-decoder models in favor of decoder-only architectures implicitly treats separate perception and production mechanisms as unnecessary---yet our results suggest that even within a fully shared-parameter, decoder-only framework, prompt framing alone is sufficient to induce a functionally meaningful asymmetry between the two modes.

Second, our methodological approach---using direct token probabilities rather than metalinguistic prompting for examination of distinction between language production and perception---proved to be successful. \citet{hu-levy-2023-prompting} demonstrated that prompted responses systematically diverge from a model's underlying probability distributions, and that this divergence increases as the prompt moves further from direct next-word prediction. This means that if we had asked the model to actually produce a rating, we would have been measuring not just how the model represents the text, but also how well it can talk about language, which is a separate ability. By using the perception prompt only as a framing context, and never actually collecting the rating, we avoid this problem. This way, we get a cleaner look at how the prompt framing itself affects the model's internal probability distributions. We suggest our methodological approach for future examinations of production and perception in LLMs.

\section{Limitations}

Several limitations should be noted. While we replicated the effect across five models (Section~\ref{sec:results:models}), all are open-weight models in the 7--9B range, and we examined only a single creative domain (simple poems), which limits generalizability. Future work should test whether the observed divergence replicates across larger models, additional model families, and other text genres. Additionally, \citet{kuribayashi2025largelanguagemodelshumanlike} argue that probabilities derived from internal layers of LLMs may better capture certain aspects of language processing than final-layer probabilities. Extending our paradigm to internal-layer probability extraction could reveal whether the production--perception divergence we observe is uniform across layers or concentrated at particular depths---a question with important implications for understanding where and how prompt framing is encoded in the model. Lastly, while we observed a difference between production and production--perception conditions, the correlation between production and perception token probabilities was high. This raises an important interpretive question: although we can establish that a statistically reliable difference exists, it remains unclear how meaningful this difference is in magnitude, and consequently, how directly these findings can be mapped onto human behavior. Future work should develop clearer criteria for what degree of divergence between conditions would constitute a theoretically meaningful difference when using LLMs as proxies for human cognitive processing.

\section*{Acknowledgments}

\subsection*{Funding}
Jiří Milička was supported by Czech Science Foundation Grant No. 24-11725S, \url{gacr.cz} (``Large language models through the prism of corpus linguistics'').
Anna Marklová was supported by Primus Grant PRIMUS/25/SSH/010.
Generating and analyzing of several datasets was supported by the project ``Human‐centred AI for a Sustainable and Adaptive Society'' (reg. no.: CZ.02.01.01/00/23\_025/0008691), co‐funded by the European Union.
\subsection*{Declaration on using AI}
We consulted Claude 4.6 Sonnet and Opus, Gemini 3 Pro, and GPT 5.2 for the stylization and editing of this article (however, all literature research was conducted by humans, and the core ideas originate entirely from the authors). The analysis scripts were also largely generated using these models (with all code thoroughly reviewed and verified by the authors).

\section{Data Availability}
The data and scripts used in this study are publicly available at \url{https://osf.io/sc3ra/overview?view_only=7a0f407eccd94f289e3506727a4430c9} (anonymous link).

\bibliography{bibliography}

\onecolumn
\appendix
\section{Per-scenario results for \texttt{Llama-3.1-8B}}
\label{app:scenario}

\begin{table*}[ht]
\centering
\small
\setlength{\tabcolsep}{4pt}
\caption{Per-scenario summary statistics for \texttt{Llama-3.1-8B}.
         Each row corresponds to one prompt pair.
         Correlation denotes per-poem Pearson~$r$; Distance stands for per-poem mean absolute distance;
         $\hat{b}$, $\hat{c}$ are exponential-decay parameters
         $y = a\,e^{-bx}+c$ fitted over positions~10--100.
         95\% bootstrap confidence intervals in brackets.}
\label{tab:scenario_results}
\begin{tabular}{lllll}
\toprule
\textbf{Scenario} & \textbf{Correlation [95\% CI]} & \textbf{Distance [95\% CI]} & \textbf{Param. $\hat{b}$ [95\% CI]} & \textbf{Param. $\hat{c}$ [95\% CI]} \\
\midrule
\multicolumn{5}{l}{\textit{Production--Perception (\textsc{prod-perc})}} \\
\textit{prod$_0$ / perc$_0$} & 0.978 [0.976, 0.979] & 0.034 [0.033, 0.035] & 0.021 [0.008, 0.036] & 0.018 [0.005, 0.024] \\
\textit{prod$_0$ / perc$_1$} & 0.974 [0.972, 0.975] & 0.038 [0.037, 0.040] & 0.010 [0.008, 0.019] & 0.004 [0.000, 0.018] \\
\textit{prod$_0$ / perc$_2$} & 0.975 [0.973, 0.976] & 0.038 [0.037, 0.039] & 0.009 [0.006, 0.019] & 0.005 [0.000, 0.021] \\
\textit{prod$_1$ / perc$_0$} & 0.974 [0.972, 0.976] & 0.037 [0.036, 0.038] & 0.009 [0.005, 0.022] & 0.007 [0.000, 0.024] \\
\textit{prod$_1$ / perc$_1$} & 0.973 [0.971, 0.974] & 0.040 [0.039, 0.041] & 0.018 [0.007, 0.034] & 0.019 [0.000, 0.028] \\
\textit{prod$_1$ / perc$_2$} & 0.973 [0.970, 0.975] & 0.039 [0.038, 0.040] & 0.026 [0.009, 0.049] & 0.025 [0.011, 0.031] \\
\textit{prod$_2$ / perc$_0$} & 0.980 [0.978, 0.981] & 0.034 [0.033, 0.035] & 0.041 [0.031, 0.051] & 0.021 [0.019, 0.023] \\
\textit{prod$_2$ / perc$_1$} & 0.977 [0.975, 0.978] & 0.037 [0.036, 0.038] & 0.049 [0.039, 0.060] & 0.025 [0.023, 0.026] \\
\textit{prod$_2$ / perc$_2$} & 0.977 [0.976, 0.979] & 0.037 [0.036, 0.038] & 0.038 [0.026, 0.050] & 0.024 [0.021, 0.026] \\
\textit{prod$_3$ / perc$_0$} & 0.982 [0.980, 0.983] & 0.028 [0.027, 0.029] & 0.057 [0.032, 0.089] & 0.020 [0.018, 0.022] \\
\textit{prod$_3$ / perc$_1$} & 0.976 [0.974, 0.978] & 0.032 [0.031, 0.033] & 0.051 [0.027, 0.080] & 0.023 [0.020, 0.026] \\
\textit{prod$_3$ / perc$_2$} & 0.978 [0.974, 0.981] & 0.030 [0.029, 0.031] & 0.043 [0.005, 0.109] & 0.020 [0.000, 0.026] \\
\midrule
\multicolumn{5}{l}{\textit{Production--Production (\textsc{prod-prod})}} \\
\textit{prod$_0$ / prod$_1$} & 0.998 [0.997, 0.998] & 0.010 [0.010, 0.010] & 0.040 [0.005, 0.170] & 0.007 [0.001, 0.009] \\
\textit{prod$_0$ / prod$_2$} & 0.990 [0.990, 0.991] & 0.023 [0.022, 0.023] & 0.020 [0.006, 0.036] & 0.013 [0.001, 0.017] \\
\textit{prod$_0$ / prod$_3$} & 0.990 [0.987, 0.991] & 0.021 [0.020, 0.022] & 0.021 [0.007, 0.038] & 0.012 [0.000, 0.015] \\
\textit{prod$_1$ / prod$_0$} & 0.996 [0.995, 0.997] & 0.011 [0.011, 0.012] & 0.263 [0.001, 0.703] & 0.008 [0.000, 0.011] \\
\textit{prod$_1$ / prod$_2$} & 0.988 [0.987, 0.989] & 0.024 [0.023, 0.025] & 0.014 [0.004, 0.041] & 0.009 [0.000, 0.019] \\
\textit{prod$_1$ / prod$_3$} & 0.989 [0.988, 0.990] & 0.023 [0.022, 0.024] & 0.063 [0.028, 0.103] & 0.018 [0.016, 0.020] \\
\textit{prod$_2$ / prod$_0$} & 0.990 [0.989, 0.991] & 0.022 [0.021, 0.022] & 0.057 [0.041, 0.077] & 0.015 [0.014, 0.017] \\
\textit{prod$_2$ / prod$_1$} & 0.991 [0.990, 0.992] & 0.022 [0.021, 0.023] & 0.040 [0.031, 0.050] & 0.015 [0.013, 0.016] \\
\textit{prod$_2$ / prod$_3$} & 0.992 [0.990, 0.993] & 0.021 [0.020, 0.022] & 0.056 [0.040, 0.073] & 0.014 [0.013, 0.016] \\
\textit{prod$_3$ / prod$_0$} & 0.990 [0.988, 0.992] & 0.019 [0.018, 0.021] & 0.030 [0.005, 0.068] & 0.011 [0.000, 0.015] \\
\textit{prod$_3$ / prod$_1$} & 0.991 [0.989, 0.992] & 0.019 [0.018, 0.020] & 0.033 [0.005, 0.069] & 0.012 [0.000, 0.015] \\
\textit{prod$_3$ / prod$_2$} & 0.990 [0.989, 0.992] & 0.020 [0.019, 0.021] & 0.055 [0.033, 0.080] & 0.014 [0.013, 0.016] \\
\bottomrule
\end{tabular}
\end{table*}

\end{document}